\documentclass[manuscript]{acmart}

\begin{document}
\title{Context-Aware Discrimination Detection in Job Vacancies using Computational Language Models}

\author{Steven Vethman}
\authornote{Both authors contributed equally to this research.}
\email{steven.vethman@tno.nl}
\orcid{0001-5435-8671}
\affiliation{%
  \institution{TNO, Netherlands Organisation for Applied Scientific Research}
  \streetaddress{Anna van Buerenplein 1}
  \city{The Hague}
  \country{The Netherlands}
  \postcode{2496 RZ}
}

\author{Ajaya Adhikari}
\authornotemark[1]
\affiliation{%
  \institution{TNO, Netherlands Organisation for Applied Scientific Research}
  \streetaddress{Anna van Buerenplein 1}
  \city{The Hague}
  \country{The Netherlands}
  \postcode{2496 RZ}
}
\email{ajaya.adhikari@tno.nl}

\author{Maaike H.T. de Boer}
\affiliation{%
  \institution{TNO, Netherlands Organisation for Applied Scientific Research}
  \streetaddress{Anna van Buerenplein 1}
  \city{The Hague}
  \country{The Netherlands}
  \postcode{2496 RZ}
}
\email{maaike.deboer@tno.nl}

\author{Joost A.G.M. van Genabeek}
\affiliation{%
  \institution{TNO, Netherlands Organisation for Applied Scientific Research}
  \streetaddress{Anna van Buerenplein 1}
  \city{The Hague}
  \country{The Netherlands}
  \postcode{2496 RZ}
}
\email{joost.vangenabeek@tno.nl}

\author{Cor J. Veenman}
\affiliation{%
  \institution{TNO, Netherlands Organisation for Applied Scientific Research}
  \streetaddress{Anna van Buerenplein 1}
  \city{The Hague}
  \country{The Netherlands}
  \postcode{2496 RZ}
}
\affiliation{
    \institution{Leiden Institute of Computer Science (LIACS), Leiden University}
    \city{Leiden}
    \country{The Netherlands}
}
\email{c.j.veenman@liacs.leidenuniv.nl}

\renewcommand{\shortauthors}{Vethman S., Adhikari A., de Boer M.H.T., van Genabeek J.A.G.M., Veenman C.J.}

\begin{abstract}

Discriminatory job vacancies are a recognized problem with strong impact on inclusiveness and belongingness.  It is disapproved worldwide, but remains persistent. Discrimination in job vacancies can be explicit by directly referring to demographic memberships of candidates. On the other hand, more implicit forms of discrimination are also present that may not always be illegal but still  influence the diversity of applicants. 
Although there is a shift towards using implicit forms, explicit written discrimination is still present in numerous job vacancies, as was recently observed for age discrimination in the Netherlands. The studies demonstrated that the lower bound for age discrimination in job vacancies was approximated between 0.16\% and 0.24\%, while further in-depth analyses showed actual numbers rise above the conservative lower bound.
Current efforts for the detection of explicit discrimination concern the identification of job vacancies containing potentially discriminating terms such as ``young” or ``male”. 
In this way, the automatic detection of potentially discriminatory job vacancies is, however, very inefficient due to the consequent low precision: for instance ``we are a young company” or ``working with mostly male patients” are phrases that contain explicit terms, while the context shows that these do not reflect discriminatory content.

In this paper, we show how state-of-the-art machine learning based computational language models can raise precision in the detection of explicit discrimination by identifying when the potentially discriminating terms are used in a discriminatory context, indicating that the sentence is indeed discriminating. We focus on gender discrimination, which indeed suffers from low precision when filtering explicit terms. First, in collaboration with oversight authorities we created a data set for gender discrimination in job vacancies. Second, we investigated a variety of machine learning based computational language models for discriminatory context detection. Third, we explored and evaluated the capability of these models to detect unforeseen discriminating terms in context. The results show that machine learning based methods can detect explicit gender discrimination with high precision and these state-of-the-art natural language processing techniques help in finding new forms of discrimination. Accordingly, the proposed methods can substantially increase the effectiveness of oversight authorities and job search websites to detect job vacancies which are highly suspected to be discriminatory. In turn, this may steer towards a fairer labour market by lowering the discrimination experienced at the start of the recruitment process.

\end{abstract}

\ccsdesc[500]{Computing methodologies~Natural language processing}
\ccsdesc[500]{Social and professional topics~Gender}

\keywords{Discrimination, Language models, Bias, Job vacancies, Context detection, Fairness, Artificial intelligence (A.I.)}
\maketitle

\section{Introduction}
Recruiting personnel for open job positions and the search for a new job by job seekers is a two way process. 
Job seekers post their CV's on suitable websites and employers post their vacancies. 
CV's are mostly in standard formats for the respective job site, while vacancies are typically in free text form. 
Whilst writing the vacancy texts the employer has the opportunity to attract the right people for an open job position in the company. 
The employer can picture a nice image of the company, but may also show preferences for a certain type of candidate. 
In other words, favoring the vacancy towards specific job seekers, which is only allowed as long as it is directed towards objective required skills. 
Otherwise, socio-demographic properties may be specified, which is worldwide disapproved by the authorities. 
Despite the awareness of such discrimination in job vacancies, it can still be found.
Discrimination in hiring has, however, moved from being explicit to implicit \cite{whysall2018cognitive}. Still, explicit discrimination in job vacancies has not vanished: in China, gender-targeted job advertisements are still common \cite{kuhn2013gender}; but also in Western countries such as the Netherlands,
recent studies in 2017, 2018 and 2019 have estimated that the lower bound for explicit age discrimination in Dutch job vacancies is between 0.16\% and 0.24\% \cite{fokkens2018leeftijd,iszwleeftijd,fokkens2020leeftijd}. Although the estimates might seem low, it still concerns thousands of violations of the law each year. The authors of \cite{fokkens2020leeftijd} also note that their algorithm for the automated analysis is conservative, given that a manual check in 2019 of 4598 job vacancies indicated that it is likely that 200\% more occurrences of explicit age discrimination are present. That is, the lower bound is just the tip of the iceberg.  

We define \textit{explicit} discrimination as directly referring to a certain demographic membership, which is based on the definition of direct discrimination in the Dutch Equal Treatment Act \cite{AWGB}. We refer to all other types of discrimination that do not directly refer to the demographic membership  as \textit{implicit}. As such, we differentiate the explicit written form of discrimination from the implicit forms that are more subjective and, therefore, less enforceable and a grey area whether it is illegal. For example, asking for a male candidate is explicit, while describing a strong or a caring candidate has implicit gender associations that is generally not illegal but still influences the diversity of applicants through an implied sense of belonging \cite{gaucher2011evidence}.

Current straight forward efforts against explicit discrimination are using the automatic detection of vacancies that contain any term on a list of potentially discriminatory terms, called a ``forbidden list''. 
Each term on the list is formulated as a regular expression (regex), e.g. ``male(s)'', such that specified alternative forms of the same term, here the plural form ``males'', are also detected.
Each term on the forbidden list also has specific exceptions such as the term ``male'' having the exception ``female'' such that vacancies specifically advertising that ``male, female or applicants of any gender are welcome to apply”, would not flag a job advertisement as discriminatory. Flagged vacancies are inspected manually to check whether the potentially discriminating term is used in a discriminatory context. One cannot ask for a ``male'' prison guard, but one can specify the guard will work with ``male'' prisoners.
This context is not detected by use of the forbidden list approach and, as a result, job vacancies containing non-discriminatory uses of  search terms are flagged, lowering the precision of the approach. Low precision is problematic here, given that false alarms distract the users of this approach from finding the job vacancies that do discriminate. 

Parties such as the Netherlands Labour Authority (NLA) and the Dutch Employment Insurance Agency (UWV) facilitate, supervise and orientate research on the recruitment process on the job market. Job search websites also do not want to host or post vacancies containing these explicit forms of discrimination. As these parties are time and resource constrained, the low precision and resulting inefficiency of the current \emph{baseline method} of the forbidden list approach is not a feasible approach to fulfill their urgent need to detect and monitor discrimination in job vacancies.

In this work, we propose the use of Machine learning (ML), the data-driven subfield of Artificial Intelligence (A.I.), to bring higher precision and efficiency to support the detection of discrimination in job vacancies. Computational language models, resulting from ML applied on written text, have recently shown high performance in various text classification problems \cite{kowsari,altinel}. The idea is to support the baseline method to grasp the contextual nature of discrimination. Recent computational language research has especially flourished in the detection of \emph{implicit} biased language that is strongly associated with demographics (gender, ethnicity) \cite{garrido2021survey,bolukbasi2016man,caliskan2017semantics}. On the other hand, the potential of detecting \emph{explicit} discrimination with computational language models has not been investigated to the best of our knowledge. 


With the following three contributions, we will demonstrate how computational language modelling can aid in the improved detection of known and unknown explicit discrimination in job vacancies.

\subsection*{C1: Create a Data Set with Labeled Explicit Discrimination in Context}
First, we design, collect and annotate a data set in collaboration with domain experts from UWV, NLA and the Netherlands Institute for Human Rights (NIHR). 
This high quality data set contains sentences from Dutch job vacancies containing potentially discriminating terms. Domain experts have annotated each sentence with a label signifying whether the term is suspected to be used in a discriminatory context. 
The domain experts involved are employees that might use such a system in practice, ensuring that the annotated label is relevant. 

\subsection*{C2: Compare Methods for the Detection of \emph{Known} Explicit Discrimination in Context}
Second, we compare a variety of text feature extraction methods, language models and supervised machine learning models to improve the \emph{precision} of baseline method for explicit discrimination detection, i.e. using explicit terms.
We investigate three state-of-the-art feature extraction methods that vary in complexity, which are complemented with three commonly used supervised learning methods, Logistic Regression\cite{cox1958regression}, XGboost\cite{xgboost}, Random Forest\cite{ho1995random}. On one hand the potential of easier-to-implement ML that use the frequency of words (bag-of-words methods with n-grams)\cite{li2020survey} or the semantic meaning of words via embeddings (word2vec)\cite{Mikolov2013}. On the other hand, we also explore advanced ML that is specifically made to incorporate the contextual meaning of words (Bidirectional Encoder Representations from Transformers, BERT~\cite{BERT}.

\subsection*{C3: Explore the Detection of \emph{Unknown} Explicit Discrimination in Context}
Third, as the collection of search terms for gender may not be complete and new forms of discrimination may spawn over time, we investigate the ability of the methods to detect the discriminatory context of unknown potentially discriminating terms. We test this with two experimental set-ups. The first set-up aims at detecting explicit discrimination given that \emph{unknown explicit terms are used}. In other words, we know that explicit terms are in the sentence, but we do not know which. The second set-up is subtly different. In that case, the aim 
is to detect explicit discrimination given that \emph{no known explicit terms are used}. This means that we do not know whether explicit terms are present in the sentence at all. This is the most general case to find new explicit discrimination.

\vspace{0.5cm}

The lay-out of the paper is as follows. In the next section, we first elaborate on related work in labor market research on discrimination, machine learning research on textual context classification and bias in language models, and market solutions for inclusive job vacancies. 
In Section 3, we describe the methods used for the creation of the data set and the experimental design choices for detecting known and unknown forms of explicit discrimination. In Section 4, we demonstrate the results: the data set and the insights on the ability of computational language models to detect the discriminatory context of known and unknown forms of explicit gender discrimination. We finalize the paper with the concluding remarks.

\section{Related Work}
To put our efforts for context aware discrimination detection in job vacancies into context we have signalled three domains of related work. First, we position ourselves within other research on discrimination in the labor market. Second, we indicate our position in the field of machine learning on textual data, called natural language processing, with emphasis on the state-of-art in context dependent text classification and bias in language models. Third and last, we overview technological solutions offered in the market to improve the fairness of job vacancies. 

\subsection{Labor Market Research on Discrimination} 

Scientific research into discrimination in the labor market has increased rapidly in recent years \cite{baert2018hiring}. Much emphasis has been placed on finding and explaining discrimination in the recruitment and selection of new personnel for companies. The studies show that candidates can experience prejudice and discrimination at all stages of the recruitment and selection process \cite{alonso2017structured,aramburu2001adverse,hough2001determinants}.

Different methods are used to detect discrimination during the recruitment and selection process. Most studies relate to field experiments \cite{baert2018hiring,bertrand2017field,neumark2018experimental,rich2014field,zschirnt2016ethnic}, sometimes combined with vignette methods \cite{bertogg2020gender}. Relatively often, these studies focus on possible unequal treatment of candidates in job interviews and selection \cite{di2020understanding,alonso2017structured, zschirnt2016ethnic}. Aside from the methodological challenges \cite{keuschnigg2016use}, many field experiments are highly context dependent, making it difficult to draw general conclusions based on the circumstances in which discrimination occurs and the factors influencing it \cite{di2020understanding,imdorf2017understanding}.

Moreover, in most field experiments and vignette studies, relatively little attention is paid to earlier steps of the recruitment process, in particular the drafting and posting of vacancies in job advertisements \cite{di2020understanding,zschirnt2016ethnic}. Vacancy texts are generally the starting point of the recruitment process, in which companies record what the job entails and what is expected of candidates. 
The way in which a vacancy is formulated often determines the entire recruitment and selection process. If a preference for a male employee is expressed in the vacancy text, a male candidate will often ultimately be selected for the job \cite{bem1973does,gaucher2011evidence}. It is, therefore, relevant to conduct further research into discriminatory formulations of vacancies in job advertisements.

Scientific studies into discriminatory formulations of job vacancies are relatively scarce. The reason is that until recently, researchers had to manually or semi-automatically extract discriminatory wording from large amounts of dissimilar texts (see for example \cite{gaucher2011evidence}). 
This was often a labour-intensive task that required a lot of time to generate results.
In recent years, new techniques have become available that can identify multiple types of discriminatory formulations in online job vacancies on a large scale \cite{ningrum2020text}. 
In the Netherlands,  \cite{fokkens2018leeftijd,iszwleeftijd,fokkens2020leeftijd} have taken an important step by automatically detecting explicit age discrimination in all Dutch-language vacancy texts that were published on the internet in 2017, 2018 and 2019.
With the help of domain experts they compiled a list of age-related formulations that are discriminatory, similar to the search terms and previously referred ``forbidden list''. Using the method of regular expressions they estimated that the lower bound on the prevalence of explicit age discrimination ranges from 0.16\%-0.24\%.

To summarize, recent social scientific research has focused on investigating discrimination in the labor market via context-dependent field experiments which suffer from limited external validity. 
On the other hand, recent automated analyses were able to expand the scale of the analysis to vacancies from all sectors, but would not yet allow for investigating the context-dependent element of discrimination.
Our paper aims to form the bridge between these efforts and acknowledge the contextual nature of gender discrimination in job vacancies of all sectors by extending the analysis of regular expression to machine learning with context-dependent text classification.

\subsection{Context Dependent Classification}
Traditionally, text classification has been done using feature extraction \cite{li2020survey}, such as extracting the separate words (``bag of words'') or combination of words (``n-grams''), and a state-of-the-art classification model, such as an SVM \cite{hassan2017deep}. This type of feature-based text representation enables to classify text fragments using any classification method developed for structured or tabular data. These representations, however, still lose valuable information contained in the text. 
With a bag of words model the order of the words disappears, and in the n-gram model there is a sparsity problem, since many sentences are unique and a lot of training data is needed to have enough frequent n-grams to train a classifier. 

In the past decade, more sophisticated feature extraction or language modelling methods are proposed that encode text fragments in a feature space where distances correspond to semantic similarity. The most known word2vec is a group of models that produce so-called semantic embeddings \cite{Mikolov2013}. These models create word embeddings using a neural network that is trained on a large document set, such as Wikipedia, Google News or Twitter. Each word is translated to a vector, and vectors of semantically similar words are close to each other in the vector space. 
This results in the famous  association example: \emph{king - man + woman = queen}~\cite{Mikolov2013}. In general, the word2vec representation works better than the older methods, but it still does not take the context of words into account as sometimes needed to express the  meaning of a word: for example with word2vec the word \textit{bank} has one vector representation, but the meaning can be a financial bank or a river bank. This issue can be mitigated to a certain extent by using domain-specific word2vec models, but these require a lot of training data to create a reliable model. Another option to include context is concatenating vectors from words or sentences left and right of the analyzed word, such as in the Flair method \cite{akbik2019flair}.

Recently, dedicated methods have been proposed to take into account the context of the words. One of the best known context-aware language models is the BERT model \cite{BERT}, which stands for Bidirectional Encoder Representation from Transformers. Transformers are a specific type of neural network architecture that can exploit sequentially correlated data, such as written language \cite{vaswani2017attention} and speech~\cite{dong2018speech}. BERT is a type of network that is build as a transformer and creates contextual embeddings. This means that the same word gets different vectors depending on the context in which it occurs. \cite{wiedemann2019does} show on different data sets that disambiguation with BERT works better than previously proposed methods.
This is mainly established by a deep bidirectional learning part, in which the left-to-right and right-to-left context is captured simultaneously (as opposed to a separate learning method in the bidirectional LSTM model named ELMo \cite{peters2018deep}). Many different types of (pre-trained) BERT models are nowadays available, such as roBERTa \cite{liu2019roberta} which uses a more robust training approach by analysing the impact of many key hyperparameters and training data size, the lighter and faster distilBERT \cite{sanh2019distilbert}, and BERTje which is trained on Dutch data \cite{BERTje}. 
In the last years, methods such as XLnet \cite{yang2019xlnet} and the GPT series from OpenAI \cite{radford2019language,brown2020language} have been proposed. 
Although these models have a higher performance in a few tasks, they have many more parameters compared to most BERT models and therefore require more data and computational power.

Within the field of discrimination detection, it is commonly known that language models have human-like biases related to undesirable societal stereotyping \cite{garrido2021survey,caliskan2017semantics}; as for example in word2vec embeddings it was shown that the vectors representing men are closer to computer programmer and women to homemaker \cite{bolukbasi2016man}. 
Similar research has been done with BERT \cite{zhang2020hurtful}; where the bias embedded in the BERT model trained on medical notes leads to classifiers having differences in performance regarding gender, language, ethnicity, and insurance status. Other studies acknowledge the ability of BERT based models to capture biased contexts and investigate methods how to mitigate biased effects in text classification \cite{zhang2020demographics,zhang2020hurtful,dearteaga2019bias}.
In the current research the aim is not to measure bias in the models or mitigate bias while using the models, but to use these models to detect contextual bias or discriminatory language. 

\subsection{Technological Market Solutions for Inclusive Job Vacancies}
Although the challenge of contextual explicit discrimination is not yet tackled, there are many companies that promote technologies enabling more inclusive writing in vacancy texts, such as Textio (\textit{textio.com}) in the U.S. and e.g. Develop Diverse (\textit{developdiverse.com}) and CorTexter (\textit{cortexter.com}) in Europe. 

Some of these companies have based their technology directly on insights from social science studies such as Gaucher \cite{gaucher2011evidence}, others follow the indications from bias measurement in language models to highlight potentially biased wording in text that may affect the diversity of applicants. 
We highlight that these technologies are pointing to implicit biased wording. Biased wording is targeted to stimulate diversity of applicants, but it may be subjective whether it is considered discriminatory, and therefore, hard to validate or to enforce. In contrast, we target to use computational language model for the detection of explicit discrimination, which in most if not all cases is illegal. 
\section{Methods}
This section delineates our methodology to attain the annotated data set (C1) and compare different computational language models in detecting known (C2) and unknown (C3) explicit forms of discrimination. To start, we motivate and specify how we selected gender discrimination as the target and how we created the annotated data set. Thereafter, we state our design choices concerning feature extraction and classification methods. Then, we detail the experimental designs for known and unknown explicit discrimination detection in context. Finally, we specify the performance metrics that we use to evaluate the detection of explicit discrimination in vacancy texts.

\subsection{Creation of the Annotated Data Set}\label{sec:method_annotation}

For the design of the annotated data set for explicit discrimination detection, we did a preliminary analysis based on interviews with domain experts from the NLA, UWV and NIHR and an exploratory data analysis on Dutch vacancies. Based on interviews, four types of discrimination were selected for further exploration: age, gender, ethnicity and nationality. 
These types were chosen given their expected frequency and context-dependent occurrence in job vacancies, such that context-aware modeling is required. 
Given that the frequency of age discrimination based on search terms was already investigated in the Netherlands \cite{iszwleeftijd}, we focused the data exploration on the grounds of gender, ethnicity and nationality.
With the help of NLA, UWV and NIHR potentially discriminating words, phrases or terms (``search terms") were gathered for these 3 grounds. Then, to explore the results of using a baseline method for gender, ethnicity and nationality, we applied the search terms to 2.4 million scraped Dutch vacancies posted on public websites in 2018.
A small random sample of 100 sentences of job vacancies for each search term was inspected by the authors to get an indication of the precision of the baseline method. We observed that for the grounds ethnicity and nationality, precision was already high. To illustrate, the use of the terms ``without an accent'' (100\%) and ``mother tongue'' (72\%) was most often unambiguously discriminating to those that had acquired proficient speech but were not born speaking the language required. For gender discrimination, we observed lower precision for terms such as ``female'' (22\%), ``male''(19\%), ``woman/women''(8\%), ``man/men''(4\%). We noticed a pattern for terms such as ``woman'': for distinguishing the discriminatory context it is important whether the word directly applies to the candidate (``looking for talented saleswomen'') or the other aspects of the job opening (``providing expert advise to women interested in our product''). 
Explicit age discrimination in job vacancies in 2018 had only three search terms with more than 300 occurrences found, with the predominant form of potential discrimination (6099 of the 7836 flagged vacancies) related to the request of a ``young'' candidate \cite{iszwleeftijd}. As gender discrimination showed more variety than age discrimination with six search terms having a frequency above 500, and discrimination on the basis of nationality and ethnicity showed higher precision than gender with the baseline method, gender discrimination in job vacancies was selected as the scope of this research. 

 In collaboration with NIHR, UWV and NLA the scope of the data collection and annotation was further specified. Annotation of gender discrimination is done related to known potentially discriminating terms, the before mentioned ``search terms'', that in combination with regular expressions form the baseline method. These terms are collected with domain experts from NIHR, UWV and NLA and stem from their experience or court cases. 
 Furthermore, the experts annotated respectively with \textit{Highly Suspected Discrimination} (HSD) or not \textit{certain} discrimination, because only courts can provide an definite judgement whether discrimination occurred. The potential for the use of computational language models is, therefore, in providing suggestions of vacancies that are probable or susceptible for discrimination. It is up to the human in the end to judge the severity and take actions. Moreover, the annotation is done at the sentence level, given the trade-off between amount of context available, i.e. text around the search term, and the speed of annotation, i.e. longer texts take longer to read and annotate.
 
 Two experts from NLA and three experts of UWV were selected (two women, three men) to annotate close to 6000 sentences of Dutch job vacancies that contained the explicit search terms. The possible labels were ``yes, I highly suspect the sentence to be discriminatory'', ``no, I do not highly suspect the sentence to be discriminatory'', or ``?: I do not know or would take longer than 30 seconds to decide''. The annotators received a general instruction together with an expert from NIHR for any questions beforehand. Afterwards, all five annotators provided their annotation independently, without the possibility of influencing each others' decisions.
 To determine the inter-annotator agreement between the five annotators an identical subset of 600 sentences (from the 6000) were given to each of the five annotators. The chosen measure for inter-annotator agreement is the Fleiss' Kappa \cite{fleiss1971measuring}. The Fleiss' Kappa extends the Cohen's Kappa and is the measure for evaluating the level of agreement between two or more annotators, when the annotation is measured on a categorical scale. 
 The 600 sentences subset is randomly selected and stratified for the different search terms to ensure that each is represented evenly. The remaining 5400 sentences were spread between the five annotators and also stratified with respect to to the search terms.

\subsection{Design choices and methods to detect explicit discrimination in context}\label{sec:methods:ai_models}
To gauge the potential of computational language models for decision support in context aware discrimination detection, a variety of feature extraction methods and machine learning models are compared with each other and the baseline method.

The models consist of three types of feature extraction methods: Bag of Words (BoW) \cite{li2020survey}, Word2Vec (W2V) \cite{Mikolov2013} and BERT \cite{BERT}.
For the BoW method, we use 1-gram and 2-gram to limit the explosion of amount of features, which we implemented with Python package Scikit-learn 0.24.2. 
For the W2V method, we averaged the vectors of the words in the sentence  \cite{henry2018vector} using Python package Spacy 3.1.1.
For the BoW and W2V approaches an additional ML classification model is needed.
Three common and state of the art types of machine learning models were used from the Scikit-learn 0.24.2 library \cite{scikit-learn}: Logistic Regression (LR), XGBoost (XG) and Random Forest (RF). 
For BERT, the pretrained Dutch version of the BERT model, BERTje \cite{BERTje}, was used.
A BERT model can apply feature extraction as well as training specifically for a supervised learning task with an integrated logistic like (softmax) function, called \textit{finetuning}, such that an additional ML classification model is not required. 
This results in seven different computational language modelling methods.

\subsection{Experimental Design for the Detection of \emph{Known} Explicit Discrimination in Context}
For the detection of known forms of explicit gender discrimination (C2), the performance of different methods is first evaluated by a comparison aided by a grid search on hyperparameters. 
The data was split into a train (70\%) and a test set, while stratifying on the label (HSD) as well as the search terms, i.e. the proportion of discriminatory sentences is similar in the train and test set for each known form of gender discrimination.
In the train set, a 4-fold cross-validation was used to evaluate each setting of the hyperparameter search space (see Table \ref{table:hyperparameters} in Appendix A).
For BERT, we used the recommended hyperparameters search space by the authors of BERT\cite{BERT}.
The best models according to the chosen performance metrics (upcoming section \ref{sec:metrics}) are selected based on the cross-validation and performance is evaluated on the test set.  
To further analyse the potential of the different methods, we created a learning curve that evaluates the performance of the method as a function of the dataset size. This is done by evaluating the chosen models with a 10-fold cross validation where the training data set increases by a twenty point logarithmic scale sampled from the 9 training folds and tested on the same 1 fold. The proportion of the label (HSD) is stratified and similar in the training en test sets for reliable comparison.    

\subsection{Experimental Design for the Detection of \emph{Unknown} Explicit Discrimination in Context}

For the detection of unknown forms of gender discrimination, we set up two experiments.
The first experiment focuses on detecting explicit discrimination given that unknown (to the model) explicit terms are used. 
This is done by splitting the data into \textit{n} groups according to the \textit{n} search terms. Then, each group functions in turn as a test set, while the models are trained on the remaining \textit{n - 1} groups.
For example, the models are trained on data pertaining to search terms such as ``male'', ``man/men'',``woman/women'', etc., but excluding the data pertaining to search term ``female'' and is consequently tested on detecting the discriminatory context in sentences with the search term ``female''. The model can exploit that explicit terms are present, but is not familiar with it. 
Hence, we evaluate the models with this leave-one term out approach to observe how well they can generalize detecting the difference between a discriminatory and a non-discriminatory context to unforeseen explicit potentially discriminating terms.

The second experiment aims to detect explicit discrimination given that no known explicit terms are used. In this set-up, we applied the models on the remaining job vacancy sentences that do not contain any of the search terms on which the labeled data set was built.
Subsequently, per computational language model we labelled the 100 most probable sentences containing explicit discrimination in context whether they are indeed highly suspected to be discriminatory. 
This exploration was done manually by the authors using commonsense.

\subsection{Metrics}\label{sec:metrics}
As described in the Introduction, the baseline method concerning the ``forbidden list'' approach has low \textit{precision}, i.e. job vacancies flagged as discriminatory are often non-discriminatory. Next to that, parties monitoring discrimination would aim for most if not all discriminatory job vacancies to be detected by the method, i.e. high \textit{recall}. We evaluated our computational language models with these measures through the precision recall curve. 
The precision recall curve reflects the trade-off of between precision and recall for all possible decision thresholds of the computational language models.  
In this graph, the user can find a desired and possible precision recall balance and choose the corresponding threshold given their resources.
For example, in scenarios with low resources a high threshold can be chosen, as this often results in high precision, hence, few but often correctly flagged job vacancies.
We choose the area under the precision recall curve or Average Precision (AP) metric to evaluate and compare the performance of different ML models because it represents a summary of all possible combinations of precision and recall values.
The area under the precision recall curve is approximated by taking the weighted mean of the precision at \textit{n} thresholds, in which the increase in recall from the previous threshold is used as weight:

\begin{equation}
\begin{aligned}
AP = \sum_{i=1}^{n}(R_{i}-R_{i-1})P_{i} \\ with \;\;
P_{i}=\frac{True\_Positive_{i}}{True\_Positive_{i}+False\_Positive_{i}} &&
R_{i}=\frac{True\_Positive_{i}}{True\_Positive_{i}+False\_Negative_{i}}
\end{aligned}
\label{eq:AP}
\end{equation}

Although the use case of discrimination detection shows high relevance for the AP metric, we do not want to rely on one metric. Therefore, in the comparison of methods for the detection of known explicit discrimination in context, we complement the AP with another metric, the Area Under the ROC (AUC). The receiver operating characteristic Curve (ROC) shows the trade-off between the false positive rate and the true positive rate. The area under this curve is another common and robust threshold independent measure to compare different ML models\cite{ling2003auc}. For the comparison of the computational language models, the AP and AUC metrics were used to both select the best hyperparameters and to evaluate the performance on the test-set. For hyperparameter optimization, we applied a standard gridsearch procedure.

\section{Results}
In this section, we present the results obtained from the described experiments for the three contributions formulated in this research. We start by demonstrating the annotated data (C1). Next, we demonstrate the performance of the machine learning methods for known forms of explicit discrimination (C2). We close by showing the results on the ability to detect unknown forms of explicit gender discrimination (C3).  

\subsection{Annotated Data}
This section elaborates on the different characteristics of the annotated data together with the inter-annotator agreement analysis.

The list of search terms for gender discrimination that are used to create the annotation data set is shown along with an English translation in the first and second columns of Table \ref{tab:contribution1}.
Note that multiple formulations and search terms are grouped in search terms \emph{other} and \emph{informal} given their infrequency.
The third and fourth column indicate the amount of sentences found for each search term and the percentage of them that are annotated with the label of HSD, respectively.
\emph{Jongen(s)} or boy(s) in English has the lowest percentage of HSD (10.9\%), while \emph{vrouwelijk(e)}
(female) and \emph{mannelijke(e)} (male) have the highest: 60.7\% and 47.1\%.
In total, 5947 sentences are annotated of which 28.8\% received the label of HSD. A precision of 28.8 \% is therefore also the performance metric for the baseline method.

Moreover, the Fleiss' Kappa showed that in general there was a moderate agreement between the annotators: Kappa =.557 (95\% CI, .565 to .611 and p-value < 0.0005). The agreement was lower for the label concerning ``?'', Kappa = 0.038, p-value = 0.003, while for the labels of interest: ``yes'' and ``no'' the Kappa values were 0.602 and 0.657 with p-value < 0.0005 signalling good agreement. 
These interpretations of moderate (0.40 - 0.60) and good agreement (0.60 - 0.80) are according to the interpretation guidelines of Altman\cite{altman1990}. 
For the purpose of detecting highly \emph{suspected} discrimination in context instead of \emph{certain} discrimination, we deem this level of agreement as of sufficient quality to pool the other 90\% of the 6000 sentences that were only annotated by one of the five annotators. For any sentence in the subset of 600 sentences where the provided labels of the annotators differ, the label is appointed by majority vote. 51 sentences received the label ``?'' and were excluded from the data for subsequent experiments. For each search term, this accounted for at most 2\% of the sentences.
\begin{table}
\centering
\begin{tabular}{llcc}
\hline
\textbf{Dutch Search Terms}                                                                                                                                                         & \textbf{English Translation}                                                                                                                                                                            & \multicolumn{1}{l}{\textbf{Frequency}} & \textbf{Percentage HSD}  \\
\hline
jongen(s)                                                                                                                                                                           & boy(s)                                                                                                                                                                                                  & 997                                    & 10.9\%                   \\
man(nen)                                                                                                                                                                            & man/men                                                                                                                                                                                                 & 997                                    & 25.6\%                   \\
mannelijk(e)                                                                                                                                                                        & male                                                                                                                                                                                                    & 505                                    & 47.1\%                   \\
dame(s)                                                                                                                                                                             & lady/ladies                                                                                                                                                                                             & 985                                    & 18.1\%                   \\
vrouw(en)                                                                                                                                                                           & woman/women                                                                                                                                                                                             & 993                                    & 20.0\%                   \\
vrouwelijk(e)                                                                                                                                                                       & female                                                                                                                                                                                                  & 1059                                   & 60.7\%                   \\
\begin{tabular}[c]{@{}l@{}}other: \\ ~~~ ben jij onze man,\\ ~~~ met ballen,\\ ~~~ enthousiaste jongen(s),\\ ~~~ enthousiaste meid(en)/meisje(s), \\ ~~~ jonge god(in)\end{tabular} & \begin{tabular}[c]{@{}l@{}}other: \\ ~~~ you are our guy, \\ ~~~ literally "with balls" means "with grit", \\ ~~~ enthusiastic boy(s), \\ ~~~ enthusiastic girl(s), \\ ~~~ young god(dess)\end{tabular} & 191                                    & 28.3\%                   \\
\begin{tabular}[c]{@{}l@{}}informal: \\ ~~~ kerel,\\ ~~~ griet,\\ ~~~ vent, \\ ~~~ gozer\end{tabular}                                                                               & \begin{tabular}[c]{@{}l@{}}informal:\\ ~~~ dude,\\ ~~~ gall, \\ ~~~ guy, \\ ~~~ dude\end{tabular}                                                                                                       & 220                                    & 17.7\%                  \\
\hline
Total & &5947 & 28.8\% \\
\hline
\end{tabular}
\caption{Description of the collected search terms used to create the annotated data set. The amount of sentences per search term are shown as well as the proportion receiving the annotation that the sentence is highly suspected to contain explicit discrimination.}
\label{tab:contribution1}
\end{table}

\subsection{Detection of \emph{Known} Explicit Discrimination in Context}
Table \ref{table:grid_search} shows the performance of the different models on the test-set.
The models based on bag of words show high scores above 72\% AP.
This indicates that just the occurrence of words in a sentence already has value in predicting whether a gender related term is used in a discriminatory way.
Word2vec models perform worse than bag of words with AP values lower than 65\%.
Bert outperforms other models with an AP of 83.3\%, which is 6.5\% higher than the best model of Bag of words.
The order of the models according to the AUC performance is the same as the AP performance.
In the rest of the experiments the best hyperparameters according to the AP metric are used for the seven types of models.

\begin{table}[!htbp]
\caption{Performance of all models on the test-set}
\label{table:grid_search}
\centering
\begin{tabular}{llrr}
    \toprule
    Vectorizer&Model&AP&AUC\\
    \midrule
             & LR & 74,0\% & 88,5\%  \\
BoW & XG            & 76,8\% & 88,7\%  \\
             & RF        & 72,4\% & 87,1\%  \\
             \hline
             & LR & 64,9\% & 82,4\%  \\
W2V     & XG            & 61,8\% & 79,7\%  \\
             & RF       & 56,6\% & 78,5\%  \\
             \hline
BERT         & BERT                & \textbf{83,3}\% & \textbf{92,6}\% \\
  \bottomrule
\end{tabular}
\end{table}

\begin{figure}[h]
  \centering
  \includegraphics[width=0.6\textwidth]{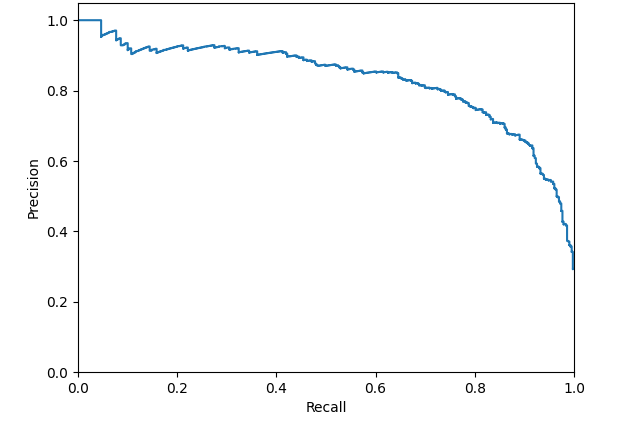}
  \caption{Precision-recall curve of the best Bert model}
  \Description{The precision value stays above 80\% between 0\% and 73\% recall. After that there is a larger decline with the lowest precision value being 29\% for the recall value of 100\%. }
  \label{fig:precision_recall_curve}
\end{figure}

The precision-recall curve of the best model (BERT) is shown in Fig. \ref{fig:precision_recall_curve}. 
The precision value stays steady above 80\% between 0\% and 73\% recall. 
After that there is a larger decline approaching the precision value of the baseline method. 
The baseline method would have a precision of 28.8\% and depending on the time and resource constraint determining how many sentences can be checked, a recall of up to 100\%. 
This graph shows that with ML there are alternatives, e.g. if a job vacancy website wants high recall of at least 80\%, the suggestions from the model would be 75\% precise.

\begin{figure}[h]
  \centering
  \includegraphics[width=\linewidth]{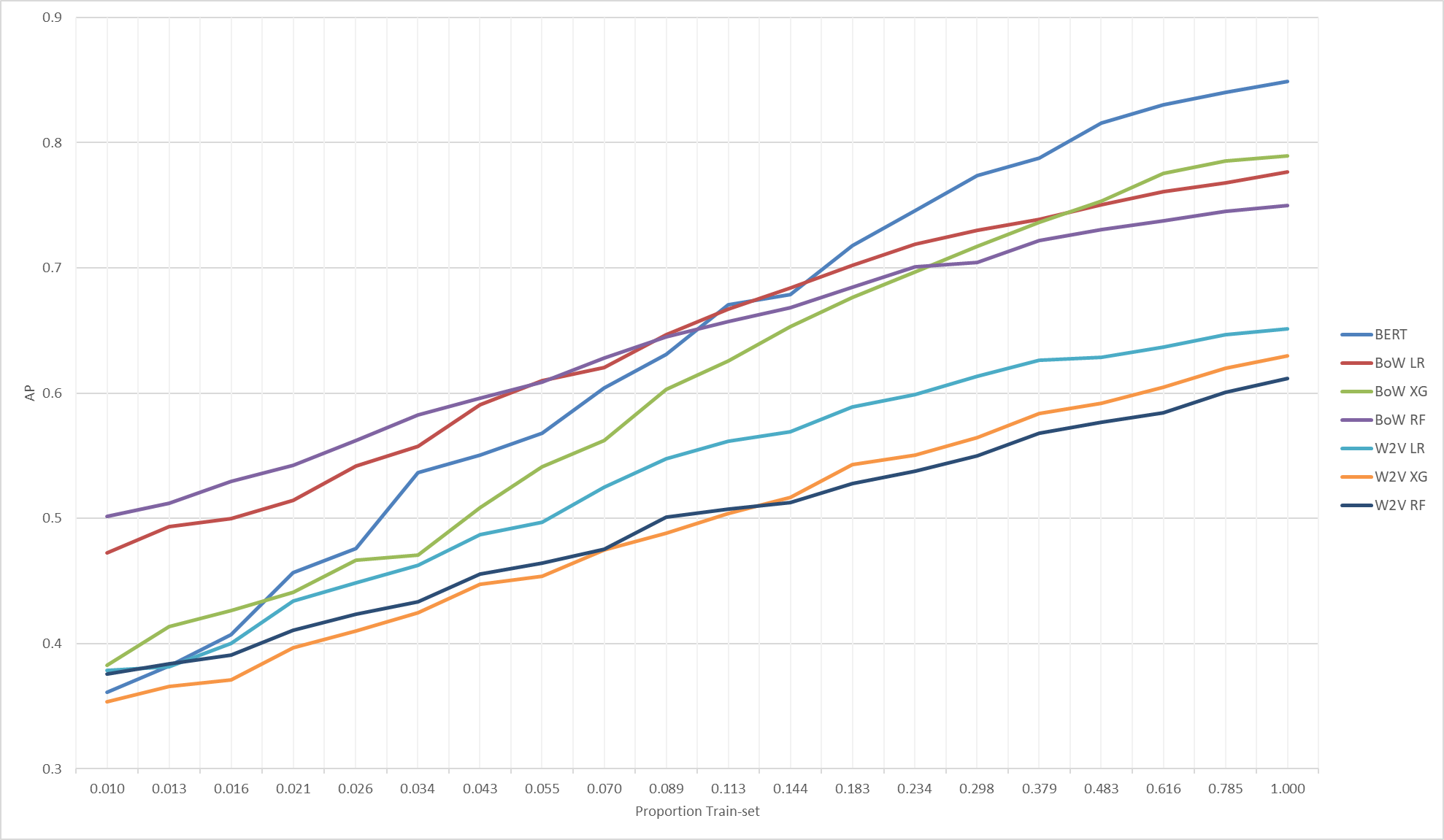}
  \caption{Learning-curve of the seven models: the x-axis shows the proportion of the train-set that was used to train the models and the y-axis the AP value on the test-set. The proportions are chosen in a logarithmic scale from 0 to 1.}
  \label{fig:learning_curve}
\end{figure}

The learning curve of the seven models are shown in figure \ref{fig:learning_curve}.
In Figure \ref{fig:learning_curve}, the learning curve indicates the performance of the methods for varying size of training data.  
We see a first cluster in the BoW LR \& RF models, that are the preferred models in case of smaller training set, up to 10\% of the data (\~600 data points). Second, the models based on language models with word2vec show to perform worse regardless of the training set size. A third cluster, BoW XG \& BERT has the most advantage of increasing amount of training data, they start with similar performance as the word2vec models, but around 18\% for BERT and 50\% for BoW XG they start outperforming the first cluster. As indicated by the grid search results of Table \ref{table:grid_search}, BERT is the preferred model for this task with BoW XG obtaining the second best score. By seeing the pattern of clusters in Figure \ref{fig:learning_curve}, it is not unlikely that the gap between them and the other models will increase with more data.   

\subsection{Detection of \emph{Unknown} Explicit Discrimination in Context}
Table \ref{tabref:searchterms} shows that BERT outperforms the other models in detecting discriminatory context in each scenario where one of the known search terms is treated as an unforeseen potentially discriminating term. The AP scores for the adjectives ``mannelijk(e)'' (0.90) and ``vrouwelijk(e)'' (0.93) stand out. 

Although these categories already had relatively high precision with base rate in the training data being 47.1\% (``mannelijk(e)'') and 60.7\% ``vrouwelijk(e)'', BERT shows great potential in surpassing this baseline in finding these and similar search terms with high precision.

Table \ref{tabref:newdiscrimination} illustrates our exploration of the methods' capabilities to detect new forms of discrimination. It is notable that BERT's 100 first suggestions for discriminatory sentences contained 37 sentences that the authors would annotate as highly susceptible of discriminatory context (HSD). From the top 10 suggestions by BERT, 7 were HSD. The computational language model based on BoW XG comes second, with 7 sentences out of 100 that are given the label of HSD.  
We note that all new forms found either concerned occupations that ended with the syllable ``man" (\textit{column M}), which contains the same meaning in Dutch as English, or concerned a sentence where explicitly only the female denomination of an occupation was requested (\textit{column F}). Occurrences are shown in the two last columns, where an asterisk (*) denotes that the gendered association of the Dutch formulation is lost in translation in the English equivalent, e.g. ``medewerkster" the female-only formulation for ``employee''.  
Furthermore, it was apparent that the classifiers and mainly BERT and BoW XG had flagged sentences mentioning occupations that are either dominated by men or dominated by women (``machine operator, plumber, forklift driver, nurse, kindergarten teacher''). 


\begin{table}
\centering
\begin{tabular}{lcccccccc}
\hline
Method & male & man/men & informal & female & lady/ladies & woman/women & boy(s) & other  \\ 
\hline
BoW LR & 0.74         & 0.55     & 0.50      & 0.84          & 0.44    & 0.43      & 0.38      & 0.50   \\
BoW RF & 0.77         & 0.59     & 0.39      & 0.84          & 0.40    & 0.39      & 0.35      & 0.51   \\
BoW XG & 0.73         & 0.59     & 0.56      & 0.82          & 0.47    & 0.41      & 0.35      & 0.44   \\
W2V LR & 0.71         & 0.58     & 0.33      & 0.79          & 0.37    & 0.34      & 0.31      & 0.49   \\
W2V RF & 0.67         & 0.52     & 0.31      & 0.78          & 0.35    & 0.39      & 0.31      & 0.52   \\
W2V XG & 0.73         & 0.52     & 0.36      & 0.80          & 0.35    & 0.39      & 0.31      & 0.50   \\
BERT   & 0.90         & 0.64     & 0.61      & 0.93          & 0.67    & 0.53      & 0.53      & 0.67   \\
\hline
\end{tabular}
\caption{AP scores for detection of unforeseen potentially discriminating search terms. The definition of the search terms \emph{other} and \emph{informal} as well as their Dutch equivalents are found in section \ref{sec:method_annotation}. }
\label{tabref:searchterms} 
\end{table} 
    
\begin{table}
\centering
\begin{tabular}{llllll} 
\hline
Method & HSD & M & F, & Occurences M                      & Occurences F                                                                 \\ 
\hline
BoW LR & 6   & 5         & 1                  & foreman, carpenter*         & doctors assistant*                                                                         \\
BoW RF & 2   & 1         & 1                  & craftsman                               & dental hygienist*                                                                           \\
BoW XG & 7   & 2         & 5                  & foreman, carpenter*                    & secretary*,  saleswoman, 
receptionist*, (...) \\
W2V LR & 0   & 0         & 0                  & -                                  & -                                                                                      \\
W2V RF & 3   & 3         & 0                  & foreman                    & -                                                                                       \\
W2V XG & 1   & 0         & 1                  & -                                  & intern*                                                                                  \\
BERT   & 37  & 25        & 9                  & carpenter*, craftsman, night guard*, (...) & employee*, assistant*, hygienist*, (...)  \\
\hline
\end{tabular}
\caption{Capability of  methods to find unknown forms of discrimination. For each method, from sentences not containing any known potentially discriminating term, 100 sentences with the highest score for discriminatory context were inspected by the authors. The count of sentences which are highly suspected to be discriminatory are shown (in column HSD) and are further categorized by either being an occupation ending with ``man'' (M) or that only the female format of the occupation was requested (F). Examples of both types are denoted in the Occurrences columns by their English equivalents. When the Dutch equivalents have a more gendered connotation, an asterisk (*) is added to denote this difference that is lost in translation. (...) is used to indicate that more than three different forms were found for a certain category.}
\label{tabref:newdiscrimination}
\end{table} 
    
\section{Concluding Remarks}
In this paper, we proposed the use of computational language models to improve the efficiency of supervision on explicit discrimination in job vacancies. First, together with domain experts, we created an annotated data set of sentences in job vacancies that are highly suspected to be discriminatory (HSD). The baseline method currently in practice for detecting discrimination obtained a precision of 28.8\% by targeting known terms that are potentially discriminating. Second, we designed seven methods that are aimed at exploiting the context of such explicit terms in order to determine whether the term is used in a discriminatory context. The methods are based on three different ways of modeling language in numerical features together with three state-of-the-art machine learning methods to predict discriminatory context. The language models are based on: bag-of-words model, word2vec model, and the BERT transformer model. The machine learning methods are logistic regression, random-forest and extreme gradient boosting (XG boost), while the BERT model has an integrated logistic like (softmax) function.

We compared the methods on their effectiveness in detecting known forms of explicit discrimination using average precision and AUC. The BERT language model clearly stands out. It has 6\% higher average precision and 4\% higher AUC than the second best, XG Boost with the bag-of-words model. The best bag-of-words model performance is 12\% higher in average precision and 9\% higher in AUC than the best word2vec performance (both with XG Boost classifier). Results indicate that for targeting the known explicit forms of gender discrimination, computational language models can  detect 80\% of the targets with 75\% precision, which increases efficiency strongly compared to the current practice  of going through all sentences with 28.8\% precision. Additionally, we evaluated the machine learning methods on their ability to detect new forms of explicit gender discrimination. Results indicate that BERT substantially outperforms the other methods in both detecting the discriminatory context related to unknown potentially discriminating terms as well as finding new potentially discriminating terms that we had not gathered together with domain experts. 

In conclusion, machine learning based computational language models have shown great potential for improving current practices of detecting explicit discrimination. The method BERT, that can capture contextual meaning of language, has shown the most promising results to aid parties such as oversight authorities and job search websites to combat against discrimination in job vacancies.

\subsection{Discussion}
We have recognized additional insights and opportunities concerning the current research.
First, it turns out that the word2vec models have relatively low performance for explicit discrimination detection, which might be due to loss of information when the word vectors of a sentence are aggregated into one vector.
Other aggregation methods such as weighting the word vectors with their tf-idf values \cite{liu2018research} or doc2vec \cite{le2014distributed} can be explored.
Second, concerning the investigation of the methods on finding new forms of explicit discrimination, we recognize that in the Dutch language the male form of occupation are often interpreted as the default and neutral form. For example, asking for a ``timmerman'' (carpenter), or ``vakman'' (craftsman) may not for everyone be considered discriminatory given that it can also be interpreted as a neutral form. However, we note that 2 out of 4 definitions of ``vakman'' (craftsman) on ``\textit{www.woorden.org/woord/vakman}'' start with the Dutch equivalent of ``a man who ... '' and we also observe the female form ``vakvrouw'' in many vacancies.
Third, we observed that the word ``enthusiastic'' was present in many of the top 100 suggested discriminatory sentences, while the sentences were actually not discriminatory. This suggest that the models have not understood that ``enthusiastic'' by itself is not an indicator for discrimination, while the word combinations ``enthusiastic boys'' and ``enthusiastic girls'' may be. 
Next to these insights and opportunities to improve the current research, we also identify two promising directions for future work.
To start, the current output of the computational language models is an estimated probability that the given sentence within a vacancy is highly suspected to be discriminatory. 
As future work, the estimated probability can be extended with explanations such as word importance such that end-users can improve their understanding of the underlying reasoning and align their level of trust accordingly.
Finally, we advise future research to repeat our work in other countries with different languages, while first identifying with domain experts which basis of discrimination is most urgent in their use case.
\begin{acks}
We thank the Netherlands Labour Authority, the Dutch Employment Insurance Agency (UWV) and the Netherlands Institute for Human Rights for their expertise and time in this collaboration.
\end{acks}

\bibliographystyle{ACM-Reference-Format}
\bibliography{ms.bib}


\begin{thebibliography}{54}


\ifx \showCODEN    \undefined \def \showCODEN     #1{\unskip}     \fi
\ifx \showDOI      \undefined \def \showDOI       #1{#1}\fi
\ifx \showISBNx    \undefined \def \showISBNx     #1{\unskip}     \fi
\ifx \showISBNxiii \undefined \def \showISBNxiii  #1{\unskip}     \fi
\ifx \showISSN     \undefined \def \showISSN      #1{\unskip}     \fi
\ifx \showLCCN     \undefined \def \showLCCN      #1{\unskip}     \fi
\ifx \shownote     \undefined \def \shownote      #1{#1}          \fi
\ifx \showarticletitle \undefined \def \showarticletitle #1{#1}   \fi
\ifx \showURL      \undefined \def \showURL       {\relax}        \fi
\providecommand\bibfield[2]{#2}
\providecommand\bibinfo[2]{#2}
\providecommand\natexlab[1]{#1}
\providecommand\showeprint[2][]{arXiv:#2}

\bibitem[\protect\citeauthoryear{??}{AWG}{2020}]%
        {AWGB}
 \bibinfo{year}{2020}\natexlab{}.
\newblock \bibinfo{title}{Algemene Wet Gelijke Behandeling (Dutch Equal
  Treatment Act)}.
\newblock
  \bibinfo{howpublished}{\url{https://wetten.overheid.nl/BWBR0006502/2020-01-01}}.
\newblock
\newblock
\shownote{Accessed: 2022-01-13}.


\bibitem[\protect\citeauthoryear{Akbik, Bergmann, Blythe, Rasul, Schweter, and
  Vollgraf}{Akbik et~al\mbox{.}}{2019}]%
        {akbik2019flair}
\bibfield{author}{\bibinfo{person}{Alan Akbik}, \bibinfo{person}{Tanja
  Bergmann}, \bibinfo{person}{Duncan Blythe}, \bibinfo{person}{Kashif Rasul},
  \bibinfo{person}{Stefan Schweter}, {and} \bibinfo{person}{Roland Vollgraf}.}
  \bibinfo{year}{2019}\natexlab{}.
\newblock \showarticletitle{FLAIR: An easy-to-use framework for
  state-of-the-art NLP}. In \bibinfo{booktitle}{\emph{Proceedings of the 2019
  Conference of the North American Chapter of the Association for Computational
  Linguistics (Demonstrations)}}. \bibinfo{pages}{54--59}.
\newblock


\bibitem[\protect\citeauthoryear{Alonso, Moscoso, and Salgado}{Alonso
  et~al\mbox{.}}{2017}]%
        {alonso2017structured}
\bibfield{author}{\bibinfo{person}{Pamela Alonso}, \bibinfo{person}{Silvia
  Moscoso}, {and} \bibinfo{person}{Jes{\'u}s~F Salgado}.}
  \bibinfo{year}{2017}\natexlab{}.
\newblock \showarticletitle{Structured behavioral interview as a legal
  guarantee for ensuring equal employment opportunities for women: A
  meta-analysis}.
\newblock \bibinfo{journal}{\emph{The European journal of psychology applied to
  legal context}} \bibinfo{volume}{9}, \bibinfo{number}{1}
  (\bibinfo{year}{2017}), \bibinfo{pages}{15--23}.
\newblock


\bibitem[\protect\citeauthoryear{Alt{\i}nel and Ganiz}{Alt{\i}nel and
  Ganiz}{2018}]%
        {altinel}
\bibfield{author}{\bibinfo{person}{Berna Alt{\i}nel} {and}
  \bibinfo{person}{Murat~Can Ganiz}.} \bibinfo{year}{2018}\natexlab{}.
\newblock \showarticletitle{Semantic text classification: A survey of past and
  recent advances}.
\newblock \bibinfo{journal}{\emph{Information Processing \& Management}}
  \bibinfo{volume}{54}, \bibinfo{number}{6} (\bibinfo{year}{2018}),
  \bibinfo{pages}{1129--1153}.
\newblock
\showISSN{0306-4573}
\urldef\tempurl%
\url{https://doi.org/10.1016/j.ipm.2018.08.001}
\showDOI{\tempurl}


\bibitem[\protect\citeauthoryear{Altman}{Altman}{1990}]%
        {altman1990}
\bibfield{author}{\bibinfo{person}{Douglas~G Altman}.}
  \bibinfo{year}{1990}\natexlab{}.
\newblock \bibinfo{booktitle}{\emph{Practical statistics for medical
  research}}.
\newblock \bibinfo{publisher}{CRC press}.
\newblock


\bibitem[\protect\citeauthoryear{Aramburu-Zabala~Higuera}{Aramburu-Zabala~Higuera}{2001}]%
        {aramburu2001adverse}
\bibfield{author}{\bibinfo{person}{Luis Aramburu-Zabala~Higuera}.}
  \bibinfo{year}{2001}\natexlab{}.
\newblock \showarticletitle{Adverse impact in personnel selection: The legal
  framework and test bias.}
\newblock \bibinfo{journal}{\emph{European Psychologist}} \bibinfo{volume}{6},
  \bibinfo{number}{2} (\bibinfo{year}{2001}), \bibinfo{pages}{103}.
\newblock


\bibitem[\protect\citeauthoryear{Baert}{Baert}{2018}]%
        {baert2018hiring}
\bibfield{author}{\bibinfo{person}{Stijn Baert}.}
  \bibinfo{year}{2018}\natexlab{}.
\newblock \showarticletitle{Hiring discrimination: An overview of (almost) all
  correspondence experiments since 2005}.
\newblock \bibinfo{journal}{\emph{Audit studies: Behind the scenes with theory,
  method, and nuance}} (\bibinfo{year}{2018}), \bibinfo{pages}{63--77}.
\newblock


\bibitem[\protect\citeauthoryear{Bem and Bem}{Bem and Bem}{1973}]%
        {bem1973does}
\bibfield{author}{\bibinfo{person}{Sandra~L Bem} {and} \bibinfo{person}{Daryl~J
  Bem}.} \bibinfo{year}{1973}\natexlab{}.
\newblock \showarticletitle{Does Sex-biased Job Advertising “Aid and Abet”
  Sex Discrimination? 1}.
\newblock \bibinfo{journal}{\emph{Journal of Applied Social Psychology}}
  \bibinfo{volume}{3}, \bibinfo{number}{1} (\bibinfo{year}{1973}),
  \bibinfo{pages}{6--18}.
\newblock


\bibitem[\protect\citeauthoryear{Bertogg, Imdorf, Hyggen, Parsanoglou, and
  Stoilova}{Bertogg et~al\mbox{.}}{2020}]%
        {bertogg2020gender}
\bibfield{author}{\bibinfo{person}{Ariane Bertogg}, \bibinfo{person}{Christian
  Imdorf}, \bibinfo{person}{Christer Hyggen}, \bibinfo{person}{Dimitris
  Parsanoglou}, {and} \bibinfo{person}{Rumiana Stoilova}.}
  \bibinfo{year}{2020}\natexlab{}.
\newblock \showarticletitle{Gender Discrimination in the Hiring of Skilled
  Professionals in Two Male-Dominated Occupational Fields: A Factorial Survey
  Experiment with Real-World Vacancies and Recruiters in Four European
  Countries}.
\newblock \bibinfo{journal}{\emph{KZfSS K{\"o}lner Zeitschrift f{\"u}r
  Soziologie und Sozialpsychologie}} \bibinfo{volume}{72}, \bibinfo{number}{1}
  (\bibinfo{year}{2020}), \bibinfo{pages}{261--289}.
\newblock


\bibitem[\protect\citeauthoryear{Bertrand and Duflo}{Bertrand and
  Duflo}{2017}]%
        {bertrand2017field}
\bibfield{author}{\bibinfo{person}{Marianne Bertrand} {and}
  \bibinfo{person}{Esther Duflo}.} \bibinfo{year}{2017}\natexlab{}.
\newblock \showarticletitle{Field experiments on discrimination}.
\newblock \bibinfo{journal}{\emph{Handbook of economic field experiments}}
  \bibinfo{volume}{1} (\bibinfo{year}{2017}), \bibinfo{pages}{309--393}.
\newblock


\bibitem[\protect\citeauthoryear{Bolukbasi, Chang, Zou, Saligrama, and
  Kalai}{Bolukbasi et~al\mbox{.}}{2016}]%
        {bolukbasi2016man}
\bibfield{author}{\bibinfo{person}{Tolga Bolukbasi}, \bibinfo{person}{Kai-Wei
  Chang}, \bibinfo{person}{James~Y Zou}, \bibinfo{person}{Venkatesh Saligrama},
  {and} \bibinfo{person}{Adam~T Kalai}.} \bibinfo{year}{2016}\natexlab{}.
\newblock \showarticletitle{Man is to computer programmer as woman is to
  homemaker? debiasing word embeddings}.
\newblock \bibinfo{journal}{\emph{Advances in neural information processing
  systems}}  \bibinfo{volume}{29} (\bibinfo{year}{2016}),
  \bibinfo{pages}{4349--4357}.
\newblock


\bibitem[\protect\citeauthoryear{Brown, Mann, Ryder, Subbiah, Kaplan, Dhariwal,
  Neelakantan, Shyam, Sastry, Askell, et~al\mbox{.}}{Brown
  et~al\mbox{.}}{2020}]%
        {brown2020language}
\bibfield{author}{\bibinfo{person}{Tom~B Brown}, \bibinfo{person}{Benjamin
  Mann}, \bibinfo{person}{Nick Ryder}, \bibinfo{person}{Melanie Subbiah},
  \bibinfo{person}{Jared Kaplan}, \bibinfo{person}{Prafulla Dhariwal},
  \bibinfo{person}{Arvind Neelakantan}, \bibinfo{person}{Pranav Shyam},
  \bibinfo{person}{Girish Sastry}, \bibinfo{person}{Amanda Askell},
  {et~al\mbox{.}}} \bibinfo{year}{2020}\natexlab{}.
\newblock \showarticletitle{Language models are few-shot learners}.
\newblock \bibinfo{journal}{\emph{arXiv preprint arXiv:2005.14165}}
  (\bibinfo{year}{2020}).
\newblock


\bibitem[\protect\citeauthoryear{Caliskan, Bryson, and Narayanan}{Caliskan
  et~al\mbox{.}}{2017}]%
        {caliskan2017semantics}
\bibfield{author}{\bibinfo{person}{Aylin Caliskan}, \bibinfo{person}{Joanna~J
  Bryson}, {and} \bibinfo{person}{Arvind Narayanan}.}
  \bibinfo{year}{2017}\natexlab{}.
\newblock \showarticletitle{Semantics derived automatically from language
  corpora contain human-like biases}.
\newblock \bibinfo{journal}{\emph{Science}} \bibinfo{volume}{356},
  \bibinfo{number}{6334} (\bibinfo{year}{2017}), \bibinfo{pages}{183--186}.
\newblock


\bibitem[\protect\citeauthoryear{Chen and Guestrin}{Chen and Guestrin}{2016}]%
        {xgboost}
\bibfield{author}{\bibinfo{person}{Tianqi Chen} {and} \bibinfo{person}{Carlos
  Guestrin}.} \bibinfo{year}{2016}\natexlab{}.
\newblock \showarticletitle{Xgboost: A scalable tree boosting system}. In
  \bibinfo{booktitle}{\emph{Proceedings of the 22nd acm sigkdd international
  conference on knowledge discovery and data mining}}.
  \bibinfo{pages}{785--794}.
\newblock


\bibitem[\protect\citeauthoryear{Cox}{Cox}{1958}]%
        {cox1958regression}
\bibfield{author}{\bibinfo{person}{David~R Cox}.}
  \bibinfo{year}{1958}\natexlab{}.
\newblock \showarticletitle{The regression analysis of binary sequences}.
\newblock \bibinfo{journal}{\emph{Journal of the Royal Statistical Society:
  Series B (Methodological)}} \bibinfo{volume}{20}, \bibinfo{number}{2}
  (\bibinfo{year}{1958}), \bibinfo{pages}{215--232}.
\newblock


\bibitem[\protect\citeauthoryear{De-Arteaga, Romanov, Wallach, Chayes, Borgs,
  Chouldechova, Geyik, Kenthapadi, and Kalai}{De-Arteaga et~al\mbox{.}}{2019}]%
        {dearteaga2019bias}
\bibfield{author}{\bibinfo{person}{Maria De-Arteaga}, \bibinfo{person}{Alexey
  Romanov}, \bibinfo{person}{Hanna Wallach}, \bibinfo{person}{Jennifer Chayes},
  \bibinfo{person}{Christian Borgs}, \bibinfo{person}{Alexandra Chouldechova},
  \bibinfo{person}{Sahin Geyik}, \bibinfo{person}{Krishnaram Kenthapadi}, {and}
  \bibinfo{person}{Adam~Tauman Kalai}.} \bibinfo{year}{2019}\natexlab{}.
\newblock \showarticletitle{Bias in bios: A case study of semantic
  representation bias in a high-stakes setting}. In
  \bibinfo{booktitle}{\emph{proceedings of the Conference on Fairness,
  Accountability, and Transparency}}. \bibinfo{pages}{120--128}.
\newblock


\bibitem[\protect\citeauthoryear{de~Vries, van Cranenburgh, Bisazza, Caselli,
  van Noord, and Nissim}{de~Vries et~al\mbox{.}}{2019}]%
        {BERTje}
\bibfield{author}{\bibinfo{person}{Wietse de Vries}, \bibinfo{person}{Andreas
  van Cranenburgh}, \bibinfo{person}{Arianna Bisazza}, \bibinfo{person}{Tommaso
  Caselli}, \bibinfo{person}{Gertjan van Noord}, {and} \bibinfo{person}{Malvina
  Nissim}.} \bibinfo{year}{2019}\natexlab{}.
\newblock \showarticletitle{Bertje: A dutch bert model}.
\newblock \bibinfo{journal}{\emph{arXiv preprint arXiv:1912.09582}}
  (\bibinfo{year}{2019}).
\newblock


\bibitem[\protect\citeauthoryear{Devlin, Chang, Lee, and Toutanova}{Devlin
  et~al\mbox{.}}{2019}]%
        {BERT}
\bibfield{author}{\bibinfo{person}{Jacob Devlin}, \bibinfo{person}{Ming-Wei
  Chang}, \bibinfo{person}{Kenton Lee}, {and} \bibinfo{person}{Kristina
  Toutanova}.} \bibinfo{year}{2019}\natexlab{}.
\newblock \showarticletitle{{BERT}: Pre-training of Deep Bidirectional
  Transformers for Language Understanding}. In
  \bibinfo{booktitle}{\emph{Proceedings of the 2019 Conference of the North
  {A}merican Chapter of the Association for Computational Linguistics: Human
  Language Technologies, Volume 1 (Long and Short Papers)}}.
  \bibinfo{publisher}{Association for Computational Linguistics},
  \bibinfo{address}{Minneapolis, Minnesota}, \bibinfo{pages}{4171--4186}.
\newblock
\urldef\tempurl%
\url{https://doi.org/10.18653/v1/N19-1423}
\showDOI{\tempurl}


\bibitem[\protect\citeauthoryear{Di~Stasio and Lancee}{Di~Stasio and
  Lancee}{2020}]%
        {di2020understanding}
\bibfield{author}{\bibinfo{person}{Valentina Di~Stasio} {and}
  \bibinfo{person}{Bram Lancee}.} \bibinfo{year}{2020}\natexlab{}.
\newblock \showarticletitle{Understanding why employers discriminate, where and
  against whom: The potential of cross-national, factorial and multi-group
  field experiments}.
\newblock \bibinfo{journal}{\emph{Research in Social Stratification and
  Mobility}}  \bibinfo{volume}{65} (\bibinfo{year}{2020}),
  \bibinfo{pages}{100463}.
\newblock


\bibitem[\protect\citeauthoryear{Dong, Xu, and Xu}{Dong et~al\mbox{.}}{2018}]%
        {dong2018speech}
\bibfield{author}{\bibinfo{person}{Linhao Dong}, \bibinfo{person}{Shuang Xu},
  {and} \bibinfo{person}{Bo Xu}.} \bibinfo{year}{2018}\natexlab{}.
\newblock \showarticletitle{Speech-Transformer: A No-Recurrence
  Sequence-to-Sequence Model for Speech Recognition}. In
  \bibinfo{booktitle}{\emph{2018 IEEE International Conference on Acoustics,
  Speech and Signal Processing (ICASSP)}}. \bibinfo{pages}{5884--5888}.
\newblock
\urldef\tempurl%
\url{https://doi.org/10.1109/ICASSP.2018.8462506}
\showDOI{\tempurl}


\bibitem[\protect\citeauthoryear{Fleiss}{Fleiss}{1971}]%
        {fleiss1971measuring}
\bibfield{author}{\bibinfo{person}{Joseph~L Fleiss}.}
  \bibinfo{year}{1971}\natexlab{}.
\newblock \showarticletitle{Measuring nominal scale agreement among many
  raters.}
\newblock \bibinfo{journal}{\emph{Psychological bulletin}}
  \bibinfo{volume}{76}, \bibinfo{number}{5} (\bibinfo{year}{1971}),
  \bibinfo{pages}{378}.
\newblock


\bibitem[\protect\citeauthoryear{Fokkens and Beukeboom}{Fokkens and
  Beukeboom}{2020}]%
        {fokkens2020leeftijd}
\bibfield{author}{\bibinfo{person}{Antske Fokkens} {and}
  \bibinfo{person}{Camiel Beukeboom}.} \bibinfo{year}{2020}\natexlab{}.
\newblock \bibinfo{booktitle}{\emph{Leeftijdsdiscriminatie in vacatureteksten:
  Een herhaalde geautomatiseerde inhoudsanalyse naar verboden
  leeftijd-gerelateerd taalgebruik in vacatureteksten uit 2017 en 2019.
  [Rapport in opdracht van het College voor de Rechten van de Mens.]}}.
\newblock


\bibitem[\protect\citeauthoryear{Fokkens, Beukeboom, and Maks}{Fokkens
  et~al\mbox{.}}{2018}]%
        {fokkens2018leeftijd}
\bibfield{author}{\bibinfo{person}{A.S. Fokkens}, \bibinfo{person}{C.J.
  Beukeboom}, {and} \bibinfo{person}{E. Maks}.}
  \bibinfo{year}{2018}\natexlab{}.
\newblock \bibinfo{booktitle}{\emph{Leeftijdsdiscriminatie in vacatureteksten:
  Een geautomatiseerde inhoudsanalyse naar verboden leeftijd-gerelateerd
  taalgebruik in vacatureteksten: Rapport in opdracht van het College voor de
  Rechten van de Mens.}}
\newblock


\bibitem[\protect\citeauthoryear{Garrido-Mu{\~n}oz, Montejo-R{\'a}ez,
  Mart{\'\i}nez-Santiago, and Ure{\~n}a-L{\'o}pez}{Garrido-Mu{\~n}oz
  et~al\mbox{.}}{2021}]%
        {garrido2021survey}
\bibfield{author}{\bibinfo{person}{Ismael Garrido-Mu{\~n}oz},
  \bibinfo{person}{Arturo Montejo-R{\'a}ez}, \bibinfo{person}{Fernando
  Mart{\'\i}nez-Santiago}, {and} \bibinfo{person}{L~Alfonso
  Ure{\~n}a-L{\'o}pez}.} \bibinfo{year}{2021}\natexlab{}.
\newblock \showarticletitle{A Survey on Bias in Deep NLP}.
\newblock \bibinfo{journal}{\emph{Applied Sciences}} \bibinfo{volume}{11},
  \bibinfo{number}{7} (\bibinfo{year}{2021}), \bibinfo{pages}{3184}.
\newblock


\bibitem[\protect\citeauthoryear{Gaucher, Friesen, and Kay}{Gaucher
  et~al\mbox{.}}{2011}]%
        {gaucher2011evidence}
\bibfield{author}{\bibinfo{person}{Danielle Gaucher}, \bibinfo{person}{Justin
  Friesen}, {and} \bibinfo{person}{Aaron~C Kay}.}
  \bibinfo{year}{2011}\natexlab{}.
\newblock \showarticletitle{Evidence that gendered wording in job
  advertisements exists and sustains gender inequality.}
\newblock \bibinfo{journal}{\emph{Journal of personality and social
  psychology}} \bibinfo{volume}{101}, \bibinfo{number}{1}
  (\bibinfo{year}{2011}), \bibinfo{pages}{109}.
\newblock


\bibitem[\protect\citeauthoryear{Hassan and Mahmood}{Hassan and
  Mahmood}{2017}]%
        {hassan2017deep}
\bibfield{author}{\bibinfo{person}{Abdalraouf Hassan} {and}
  \bibinfo{person}{Ausif Mahmood}.} \bibinfo{year}{2017}\natexlab{}.
\newblock \showarticletitle{Deep learning for sentence classification}. In
  \bibinfo{booktitle}{\emph{2017 IEEE Long Island Systems, Applications and
  Technology Conference (LISAT)}}. IEEE, \bibinfo{pages}{1--5}.
\newblock


\bibitem[\protect\citeauthoryear{Henry, Cuffy, and McInnes}{Henry
  et~al\mbox{.}}{2018}]%
        {henry2018vector}
\bibfield{author}{\bibinfo{person}{Sam Henry}, \bibinfo{person}{Clint Cuffy},
  {and} \bibinfo{person}{Bridget~T McInnes}.} \bibinfo{year}{2018}\natexlab{}.
\newblock \showarticletitle{Vector representations of multi-word terms for
  semantic relatedness}.
\newblock \bibinfo{journal}{\emph{Journal of biomedical informatics}}
  \bibinfo{volume}{77} (\bibinfo{year}{2018}), \bibinfo{pages}{111--119}.
\newblock


\bibitem[\protect\citeauthoryear{Ho}{Ho}{1995}]%
        {ho1995random}
\bibfield{author}{\bibinfo{person}{Tin~Kam Ho}.}
  \bibinfo{year}{1995}\natexlab{}.
\newblock \showarticletitle{Random decision forests}. In
  \bibinfo{booktitle}{\emph{Proceedings of 3rd international conference on
  document analysis and recognition}}, Vol.~\bibinfo{volume}{1}. IEEE,
  \bibinfo{pages}{278--282}.
\newblock


\bibitem[\protect\citeauthoryear{Hough, Oswald, and Ployhart}{Hough
  et~al\mbox{.}}{2001}]%
        {hough2001determinants}
\bibfield{author}{\bibinfo{person}{Leatta~M Hough},
  \bibinfo{person}{Frederick~L Oswald}, {and} \bibinfo{person}{Robert~E
  Ployhart}.} \bibinfo{year}{2001}\natexlab{}.
\newblock \showarticletitle{Determinants, detection and amelioration of adverse
  impact in personnel selection procedures: Issues, evidence and lessons
  learned}.
\newblock \bibinfo{journal}{\emph{International Journal of Selection and
  Assessment}} \bibinfo{volume}{9}, \bibinfo{number}{1-2}
  (\bibinfo{year}{2001}), \bibinfo{pages}{152--194}.
\newblock


\bibitem[\protect\citeauthoryear{Imdorf}{Imdorf}{2017}]%
        {imdorf2017understanding}
\bibfield{author}{\bibinfo{person}{Christian Imdorf}.}
  \bibinfo{year}{2017}\natexlab{}.
\newblock \showarticletitle{Understanding discrimination in hiring apprentices:
  how training companies use ethnicity to avoid organisational trouble}.
\newblock \bibinfo{journal}{\emph{Journal of Vocational Education \& Training}}
  \bibinfo{volume}{69}, \bibinfo{number}{3} (\bibinfo{year}{2017}),
  \bibinfo{pages}{405--423}.
\newblock


\bibitem[\protect\citeauthoryear{ISZW}{ISZW}{2020}]%
        {iszwleeftijd}
\bibfield{author}{\bibinfo{person}{ISZW}.} \bibinfo{year}{2020}\natexlab{}.
\newblock \bibinfo{title}{Leeftijdsdiscriminatie in vacatureteksten}.
\newblock
  \bibinfo{howpublished}{\url{https://www.rijksoverheid.nl/documenten/rapporten/2020/04/23/leeftijdsdiscriminatie-in-vacatureteksten}}.
\newblock


\bibitem[\protect\citeauthoryear{Keuschnigg and Wolbring}{Keuschnigg and
  Wolbring}{2016}]%
        {keuschnigg2016use}
\bibfield{author}{\bibinfo{person}{Marc Keuschnigg} {and}
  \bibinfo{person}{Tobias Wolbring}.} \bibinfo{year}{2016}\natexlab{}.
\newblock \showarticletitle{The use of field experiments to study mechanisms of
  discrimination}.
\newblock \bibinfo{journal}{\emph{Analyse \& Kritik}} \bibinfo{volume}{38},
  \bibinfo{number}{1} (\bibinfo{year}{2016}), \bibinfo{pages}{179--202}.
\newblock


\bibitem[\protect\citeauthoryear{Kowsari, Jafari~Meimandi, Heidarysafa, Mendu,
  Barnes, and Brown}{Kowsari et~al\mbox{.}}{2019}]%
        {kowsari}
\bibfield{author}{\bibinfo{person}{Kamran Kowsari}, \bibinfo{person}{Kiana
  Jafari~Meimandi}, \bibinfo{person}{Mojtaba Heidarysafa},
  \bibinfo{person}{Sanjana Mendu}, \bibinfo{person}{Laura Barnes}, {and}
  \bibinfo{person}{Donald Brown}.} \bibinfo{year}{2019}\natexlab{}.
\newblock \showarticletitle{Text Classification Algorithms: A Survey}.
\newblock \bibinfo{journal}{\emph{Information}} \bibinfo{volume}{10},
  \bibinfo{number}{4} (\bibinfo{year}{2019}).
\newblock
\showISSN{2078-2489}
\urldef\tempurl%
\url{https://doi.org/10.3390/info10040150}
\showDOI{\tempurl}


\bibitem[\protect\citeauthoryear{Kuhn and Shen}{Kuhn and Shen}{2013}]%
        {kuhn2013gender}
\bibfield{author}{\bibinfo{person}{Peter Kuhn} {and} \bibinfo{person}{Kailing
  Shen}.} \bibinfo{year}{2013}\natexlab{}.
\newblock \showarticletitle{Gender discrimination in job ads: Evidence from
  china}.
\newblock \bibinfo{journal}{\emph{The Quarterly Journal of Economics}}
  \bibinfo{volume}{128}, \bibinfo{number}{1} (\bibinfo{year}{2013}),
  \bibinfo{pages}{287--336}.
\newblock


\bibitem[\protect\citeauthoryear{Le and Mikolov}{Le and Mikolov}{2014}]%
        {le2014distributed}
\bibfield{author}{\bibinfo{person}{Quoc Le} {and} \bibinfo{person}{Tomas
  Mikolov}.} \bibinfo{year}{2014}\natexlab{}.
\newblock \showarticletitle{Distributed representations of sentences and
  documents}. In \bibinfo{booktitle}{\emph{International conference on machine
  learning}}. PMLR, \bibinfo{pages}{1188--1196}.
\newblock


\bibitem[\protect\citeauthoryear{Li, Peng, Li, Xia, Yang, Sun, Yu, and He}{Li
  et~al\mbox{.}}{2020}]%
        {li2020survey}
\bibfield{author}{\bibinfo{person}{Qian Li}, \bibinfo{person}{Hao Peng},
  \bibinfo{person}{Jianxin Li}, \bibinfo{person}{Congying Xia},
  \bibinfo{person}{Renyu Yang}, \bibinfo{person}{Lichao Sun},
  \bibinfo{person}{Philip~S Yu}, {and} \bibinfo{person}{Lifang He}.}
  \bibinfo{year}{2020}\natexlab{}.
\newblock \showarticletitle{A survey on text classification: From shallow to
  deep learning}.
\newblock \bibinfo{journal}{\emph{arXiv preprint arXiv:2008.00364}}
  (\bibinfo{year}{2020}).
\newblock


\bibitem[\protect\citeauthoryear{Ling, Huang, Zhang, et~al\mbox{.}}{Ling
  et~al\mbox{.}}{2003}]%
        {ling2003auc}
\bibfield{author}{\bibinfo{person}{Charles~X Ling}, \bibinfo{person}{Jin
  Huang}, \bibinfo{person}{Harry Zhang}, {et~al\mbox{.}}}
  \bibinfo{year}{2003}\natexlab{}.
\newblock \showarticletitle{AUC: a statistically consistent and more
  discriminating measure than accuracy}. In \bibinfo{booktitle}{\emph{Ijcai}},
  Vol.~\bibinfo{volume}{3}. \bibinfo{pages}{519--524}.
\newblock


\bibitem[\protect\citeauthoryear{Liu, Sheng, Wei, and Yang}{Liu
  et~al\mbox{.}}{2018}]%
        {liu2018research}
\bibfield{author}{\bibinfo{person}{Cai-zhi Liu}, \bibinfo{person}{Yan-xiu
  Sheng}, \bibinfo{person}{Zhi-qiang Wei}, {and} \bibinfo{person}{Yong-Quan
  Yang}.} \bibinfo{year}{2018}\natexlab{}.
\newblock \showarticletitle{Research of text classification based on improved
  TF-IDF algorithm}. In \bibinfo{booktitle}{\emph{2018 IEEE International
  Conference of Intelligent Robotic and Control Engineering (IRCE)}}. IEEE,
  \bibinfo{pages}{218--222}.
\newblock


\bibitem[\protect\citeauthoryear{Liu, Ott, Goyal, Du, Joshi, Chen, Levy, Lewis,
  Zettlemoyer, and Stoyanov}{Liu et~al\mbox{.}}{2019}]%
        {liu2019roberta}
\bibfield{author}{\bibinfo{person}{Yinhan Liu}, \bibinfo{person}{Myle Ott},
  \bibinfo{person}{Naman Goyal}, \bibinfo{person}{Jingfei Du},
  \bibinfo{person}{Mandar Joshi}, \bibinfo{person}{Danqi Chen},
  \bibinfo{person}{Omer Levy}, \bibinfo{person}{Mike Lewis},
  \bibinfo{person}{Luke Zettlemoyer}, {and} \bibinfo{person}{Veselin
  Stoyanov}.} \bibinfo{year}{2019}\natexlab{}.
\newblock \showarticletitle{Roberta: A robustly optimized bert pretraining
  approach}.
\newblock \bibinfo{journal}{\emph{arXiv preprint arXiv:1907.11692}}
  (\bibinfo{year}{2019}).
\newblock


\bibitem[\protect\citeauthoryear{Mikolov, Sutskever, Chen, Corrado, and
  Dean}{Mikolov et~al\mbox{.}}{2013}]%
        {Mikolov2013}
\bibfield{author}{\bibinfo{person}{Tomas Mikolov}, \bibinfo{person}{Ilya
  Sutskever}, \bibinfo{person}{Kai Chen}, \bibinfo{person}{Greg~S Corrado},
  {and} \bibinfo{person}{Jeff Dean}.} \bibinfo{year}{2013}\natexlab{}.
\newblock \showarticletitle{Distributed representations of words and phrases
  and their compositionality}. In \bibinfo{booktitle}{\emph{Adv. in neural
  information processing systems}}. \bibinfo{pages}{3111--3119}.
\newblock


\bibitem[\protect\citeauthoryear{Neumark}{Neumark}{2018}]%
        {neumark2018experimental}
\bibfield{author}{\bibinfo{person}{David Neumark}.}
  \bibinfo{year}{2018}\natexlab{}.
\newblock \showarticletitle{Experimental research on labor market
  discrimination}.
\newblock \bibinfo{journal}{\emph{Journal of Economic Literature}}
  \bibinfo{volume}{56}, \bibinfo{number}{3} (\bibinfo{year}{2018}),
  \bibinfo{pages}{799--866}.
\newblock


\bibitem[\protect\citeauthoryear{Ningrum, Pansombut, and Ueranantasun}{Ningrum
  et~al\mbox{.}}{2020}]%
        {ningrum2020text}
\bibfield{author}{\bibinfo{person}{Panggih~Kusuma Ningrum},
  \bibinfo{person}{Tatdow Pansombut}, {and} \bibinfo{person}{Attachai
  Ueranantasun}.} \bibinfo{year}{2020}\natexlab{}.
\newblock \showarticletitle{Text mining of online job advertisements to
  identify direct discrimination during job hunting process: A case study in
  Indonesia}.
\newblock \bibinfo{journal}{\emph{Plos one}} \bibinfo{volume}{15},
  \bibinfo{number}{6} (\bibinfo{year}{2020}), \bibinfo{pages}{e0233746}.
\newblock


\bibitem[\protect\citeauthoryear{Pedregosa, Varoquaux, Gramfort, Michel,
  Thirion, Grisel, Blondel, Prettenhofer, Weiss, Dubourg, Vanderplas, Passos,
  Cournapeau, Brucher, Perrot, and Duchesnay}{Pedregosa et~al\mbox{.}}{2011}]%
        {scikit-learn}
\bibfield{author}{\bibinfo{person}{F. Pedregosa}, \bibinfo{person}{G.
  Varoquaux}, \bibinfo{person}{A. Gramfort}, \bibinfo{person}{V. Michel},
  \bibinfo{person}{B. Thirion}, \bibinfo{person}{O. Grisel},
  \bibinfo{person}{M. Blondel}, \bibinfo{person}{P. Prettenhofer},
  \bibinfo{person}{R. Weiss}, \bibinfo{person}{V. Dubourg}, \bibinfo{person}{J.
  Vanderplas}, \bibinfo{person}{A. Passos}, \bibinfo{person}{D. Cournapeau},
  \bibinfo{person}{M. Brucher}, \bibinfo{person}{M. Perrot}, {and}
  \bibinfo{person}{E. Duchesnay}.} \bibinfo{year}{2011}\natexlab{}.
\newblock \showarticletitle{Scikit-learn: Machine Learning in {P}ython}.
\newblock \bibinfo{journal}{\emph{Journal of Machine Learning Research}}
  \bibinfo{volume}{12} (\bibinfo{year}{2011}), \bibinfo{pages}{2825--2830}.
\newblock


\bibitem[\protect\citeauthoryear{Peters, Neumann, Iyyer, Gardner, Clark, Lee,
  and Zettlemoyer}{Peters et~al\mbox{.}}{2018}]%
        {peters2018deep}
\bibfield{author}{\bibinfo{person}{Matthew~E Peters}, \bibinfo{person}{Mark
  Neumann}, \bibinfo{person}{Mohit Iyyer}, \bibinfo{person}{Matt Gardner},
  \bibinfo{person}{Christopher Clark}, \bibinfo{person}{Kenton Lee}, {and}
  \bibinfo{person}{Luke Zettlemoyer}.} \bibinfo{year}{2018}\natexlab{}.
\newblock \showarticletitle{Deep contextualized word representations}.
\newblock \bibinfo{journal}{\emph{arXiv preprint arXiv:1802.05365}}
  (\bibinfo{year}{2018}).
\newblock


\bibitem[\protect\citeauthoryear{Radford, Wu, Child, Luan, Amodei, and
  Sutskever}{Radford et~al\mbox{.}}{2019}]%
        {radford2019language}
\bibfield{author}{\bibinfo{person}{Alec Radford}, \bibinfo{person}{Jeff Wu},
  \bibinfo{person}{Rewon Child}, \bibinfo{person}{David Luan},
  \bibinfo{person}{Dario Amodei}, {and} \bibinfo{person}{Ilya Sutskever}.}
  \bibinfo{year}{2019}\natexlab{}.
\newblock \showarticletitle{Language Models are Unsupervised Multitask
  Learners}.
\newblock  (\bibinfo{year}{2019}).
\newblock


\bibitem[\protect\citeauthoryear{Rich}{Rich}{2014}]%
        {rich2014field}
\bibfield{author}{\bibinfo{person}{Judith Rich}.}
  \bibinfo{year}{2014}\natexlab{}.
\newblock \showarticletitle{What do field experiments of discrimination in
  markets tell us? A meta analysis of studies conducted since 2000}.
\newblock  (\bibinfo{year}{2014}).
\newblock


\bibitem[\protect\citeauthoryear{Sanh, Debut, Chaumond, and Wolf}{Sanh
  et~al\mbox{.}}{2019}]%
        {sanh2019distilbert}
\bibfield{author}{\bibinfo{person}{Victor Sanh}, \bibinfo{person}{Lysandre
  Debut}, \bibinfo{person}{Julien Chaumond}, {and} \bibinfo{person}{Thomas
  Wolf}.} \bibinfo{year}{2019}\natexlab{}.
\newblock \showarticletitle{DistilBERT, a distilled version of BERT: smaller,
  faster, cheaper and lighter}.
\newblock \bibinfo{journal}{\emph{arXiv preprint arXiv:1910.01108}}
  (\bibinfo{year}{2019}).
\newblock


\bibitem[\protect\citeauthoryear{Vaswani, Shazeer, Parmar, Uszkoreit, Jones,
  Gomez, Kaiser, and Polosukhin}{Vaswani et~al\mbox{.}}{2017}]%
        {vaswani2017attention}
\bibfield{author}{\bibinfo{person}{Ashish Vaswani}, \bibinfo{person}{Noam
  Shazeer}, \bibinfo{person}{Niki Parmar}, \bibinfo{person}{Jakob Uszkoreit},
  \bibinfo{person}{Llion Jones}, \bibinfo{person}{Aidan~N Gomez},
  \bibinfo{person}{{\L}ukasz Kaiser}, {and} \bibinfo{person}{Illia
  Polosukhin}.} \bibinfo{year}{2017}\natexlab{}.
\newblock \showarticletitle{Attention is all you need}. In
  \bibinfo{booktitle}{\emph{Advances in neural information processing
  systems}}. \bibinfo{pages}{5998--6008}.
\newblock


\bibitem[\protect\citeauthoryear{Whysall}{Whysall}{2018}]%
        {whysall2018cognitive}
\bibfield{author}{\bibinfo{person}{Zara Whysall}.}
  \bibinfo{year}{2018}\natexlab{}.
\newblock \showarticletitle{Cognitive biases in recruitment, selection, and
  promotion: The risk of subconscious discrimination}.
\newblock \bibinfo{journal}{\emph{Hidden inequalities in the workplace}}
  (\bibinfo{year}{2018}), \bibinfo{pages}{215--243}.
\newblock


\bibitem[\protect\citeauthoryear{Wiedemann, Remus, Chawla, and
  Biemann}{Wiedemann et~al\mbox{.}}{2019}]%
        {wiedemann2019does}
\bibfield{author}{\bibinfo{person}{Gregor Wiedemann}, \bibinfo{person}{Steffen
  Remus}, \bibinfo{person}{Avi Chawla}, {and} \bibinfo{person}{Chris Biemann}.}
  \bibinfo{year}{2019}\natexlab{}.
\newblock \showarticletitle{Does BERT make any sense? Interpretable word sense
  disambiguation with contextualized embeddings}.
\newblock \bibinfo{journal}{\emph{arXiv preprint arXiv:1909.10430}}
  (\bibinfo{year}{2019}).
\newblock


\bibitem[\protect\citeauthoryear{Yang, Dai, Yang, Carbonell, Salakhutdinov, and
  Le}{Yang et~al\mbox{.}}{2019}]%
        {yang2019xlnet}
\bibfield{author}{\bibinfo{person}{Zhilin Yang}, \bibinfo{person}{Zihang Dai},
  \bibinfo{person}{Yiming Yang}, \bibinfo{person}{Jaime Carbonell},
  \bibinfo{person}{Russ~R Salakhutdinov}, {and} \bibinfo{person}{Quoc~V Le}.}
  \bibinfo{year}{2019}\natexlab{}.
\newblock \showarticletitle{Xlnet: Generalized autoregressive pretraining for
  language understanding}.
\newblock \bibinfo{journal}{\emph{Advances in neural information processing
  systems}}  \bibinfo{volume}{32} (\bibinfo{year}{2019}).
\newblock


\bibitem[\protect\citeauthoryear{Zhang, Bai, Zhang, Bai, Zhu, and Zhao}{Zhang
  et~al\mbox{.}}{2020a}]%
        {zhang2020demographics}
\bibfield{author}{\bibinfo{person}{Guanhua Zhang}, \bibinfo{person}{Bing Bai},
  \bibinfo{person}{Junqi Zhang}, \bibinfo{person}{Kun Bai},
  \bibinfo{person}{Conghui Zhu}, {and} \bibinfo{person}{Tiejun Zhao}.}
  \bibinfo{year}{2020}\natexlab{a}.
\newblock \showarticletitle{Demographics should not be the reason of toxicity:
  Mitigating discrimination in text classifications with instance weighting}.
\newblock \bibinfo{journal}{\emph{arXiv preprint arXiv:2004.14088}}
  (\bibinfo{year}{2020}).
\newblock


\bibitem[\protect\citeauthoryear{Zhang, Lu, Abdalla, McDermott, and
  Ghassemi}{Zhang et~al\mbox{.}}{2020b}]%
        {zhang2020hurtful}
\bibfield{author}{\bibinfo{person}{Haoran Zhang}, \bibinfo{person}{Amy~X Lu},
  \bibinfo{person}{Mohamed Abdalla}, \bibinfo{person}{Matthew McDermott}, {and}
  \bibinfo{person}{Marzyeh Ghassemi}.} \bibinfo{year}{2020}\natexlab{b}.
\newblock \showarticletitle{Hurtful words: quantifying biases in clinical
  contextual word embeddings}. In \bibinfo{booktitle}{\emph{proceedings of the
  ACM Conference on Health, Inference, and Learning}}.
  \bibinfo{pages}{110--120}.
\newblock


\bibitem[\protect\citeauthoryear{Zschirnt and Ruedin}{Zschirnt and
  Ruedin}{2016}]%
        {zschirnt2016ethnic}
\bibfield{author}{\bibinfo{person}{Eva Zschirnt} {and} \bibinfo{person}{Didier
  Ruedin}.} \bibinfo{year}{2016}\natexlab{}.
\newblock \showarticletitle{Ethnic discrimination in hiring decisions: a
  meta-analysis of correspondence tests 1990--2015}.
\newblock \bibinfo{journal}{\emph{Journal of Ethnic and Migration Studies}}
  \bibinfo{volume}{42}, \bibinfo{number}{7} (\bibinfo{year}{2016}),
  \bibinfo{pages}{1115--1134}.
\newblock


\end{thebibliography}

\appendix

\section{Specification of Hyperparameters}

\begin{table}[hbtp]
\centering
\begin{tabular}{llll} 
\midrule
\textbf{Model}               & \textbf{HP Search Space}                                                                                                                                                                     & \textbf{Best BoW HP}                                                                                                                & \textbf{\textbf{Best W2V HP}}                                                                                                        \\ 
\midrule
\textbf{Logistic Regression} & \begin{tabular}[c]{@{}l@{}}C: (0.01, 0.05, 0.1 , 0.5,\\~ ~ 1, 5, 10, 50, 100) \\Solver: (lbfgs, liblinear)\\Penelty: l2\\Class weight: balanced\end{tabular}                                 & \begin{tabular}[c]{@{}l@{}}C: 0.5\\Solver: liblinear\\Penelty: l2\\Class weight:~balanced\end{tabular}                              & \begin{tabular}[c]{@{}l@{}}C: 0.1\\Solver: lbfgs\\Penelty: l2\\Class weight:~balanced\end{tabular}                                   \\ 
\midrule
\textbf{XGBoost}             & \begin{tabular}[c]{@{}l@{}}Min child weight: (2, 5, 10)\\Learning rate: (0.3, 0.2, 0.1,\\~ ~0.05, 0.01, 0.005)\\Max depth: (10, 50, 100)\\Scale positive weight: 2.5\end{tabular}            & \begin{tabular}[c]{@{}l@{}}Min child weight: 2\\Learning rate: 0.3\\Max depth: 10\\Scale positive weight: 2.5\end{tabular}          & \begin{tabular}[c]{@{}l@{}}Min child weight: 2\\Learning rate: 0.2\\Max depth: 50\\Scale positive weight: 2.5\end{tabular}           \\ 
\midrule
\textbf{Random Forest}       & \begin{tabular}[c]{@{}l@{}}Max depth: (10, 50, 100)\\Number of estimators: (200,\\~ ~600, 1000, 1400, 2000)\\Minimum samples split: (2, \\~ ~5, 10, 50)\\Class weight: balanced\end{tabular} & \begin{tabular}[c]{@{}l@{}}Max depth: 100\\Number of estimators: 200\\Minimum samples split: 5\\Class weight: balanced\end{tabular} & \begin{tabular}[c]{@{}l@{}}Max depth: 50\\Number of estimators: 1400\\Minimum samples split: 2\\Class weight: balanced\end{tabular}  \\ 
\midrule
\textbf{BERT}                & \begin{tabular}[c]{@{}l@{}}Learning rate: (2e-5, 3e-5,\\~ ~5e-5)\\Epoch amount: (2, 3, 4)\\Batch size: (16, 32)\end{tabular}                                                                 & \multicolumn{2}{l}{\begin{tabular}[c]{@{}l@{}}\textbf{Best HP:}\\Learning rate: 3e-5\\Epoch amount: 3\\Batch size: 16\end{tabular}}                                                                                                                                         \\
\midrule
\end{tabular}
\caption{The hyperparameter (HP) search space for LR, XB, RF and BERT models together with their optimal values according to the AP metric. For LR, XB and RF the hyperparameters are determined for both BoW and W2V vectorizers. The default values were chosen for all other hyperparameters that are not present in this table.}
\label{table:hyperparameters}
\end{table}

\end{document}